\pdfoutput=1

\documentclass[11pt]{article}

\usepackage[]{ACL2023}

\usepackage{times}
\usepackage{latexsym}
\usepackage{subcaption,booktabs}
\usepackage{color, colortbl}

\usepackage{algorithm}
\usepackage{algpseudocode}
\algrenewcommand\algorithmicindent{0.6em}
\usepackage{graphicx}
\usepackage{bbold}
\definecolor{bettergreen}{rgb}{0.13, 0.55, 0.13}
\definecolor{verylightgray}{gray}{0.94}
\usepackage{amsmath}
\usepackage{comment}
\usepackage{paralist}
\newcommand{\ourname}[1]{}
\renewcommand{\ourname}[1]{{\color{black} \textsc{SymbolicToM}}}

\usepackage{footnote}
\makesavenoteenv{tabular}
\makesavenoteenv{table}

\usepackage[T1]{fontenc}

\usepackage[utf8]{inputenc}

\usepackage{microtype}

\usepackage{inconsolata}

\title{Minding Language Models' (Lack of) Theory of Mind: \\
A Plug-and-Play Multi-Character Belief Tracker\vspace{0.3em}}  

\author{Melanie Sclar\textsuperscript{1} \ \ \ \ Sachin Kumar\textsuperscript{2} \ \ \ \ Peter West\textsuperscript{1} \ \ \ \ Alane Suhr\textsuperscript{3}  \ \ \ \ \\
\textbf{Yejin Choi}\textsuperscript{1,3} \ \ \ \textbf{Yulia Tsvetkov}\textsuperscript{1} \\
\textsuperscript{1}Paul G. Allen School of Computer Science \& Engineering, University of Washington \\
\textsuperscript{2}Language Technologies Institute, Carnegie Mellon University\\
\textsuperscript{3}Allen Institute for Artificial Intelligence \\
\texttt{msclar@cs.washington.edu}
}

\begin{document}
\maketitle
\begin{abstract}

Theory of Mind (ToM)---the ability to reason about the mental states of other people---is a key element of our social intelligence. Yet, despite their ever more impressive performance, large-scale neural language models still lack basic theory of mind capabilities out-of-the-box. We posit that simply scaling up models will not imbue them with theory of mind due to the inherently \emph{symbolic} and \emph{implicit} nature of the phenomenon, and instead investigate an alternative: can we design a decoding-time algorithm that enhances theory of mind of off-the-shelf neural language models without explicit supervision? We present \ourname{}, a plug-and-play approach to reason about the belief states of multiple characters in reading comprehension tasks via explicit symbolic representation. More concretely, our approach tracks each entity's beliefs, their estimation of other entities' beliefs, and higher-order levels of reasoning, all through graphical representations, allowing for more precise and interpretable reasoning than previous approaches. Empirical results on the well-known ToMi benchmark \citep{le2019revisiting} demonstrate that \ourname{} dramatically enhances off-the-shelf neural networks' theory of mind in a zero-shot setting while showing robust out-of-distribution performance compared to supervised baselines. Our work also reveals spurious patterns in existing theory of mind benchmarks, emphasizing the importance of out-of-distribution evaluation and methods that do not overfit a particular dataset.
\end{abstract}

\section{Introduction}

\begin{figure}[t]
    \centering
\includegraphics[width=\linewidth]{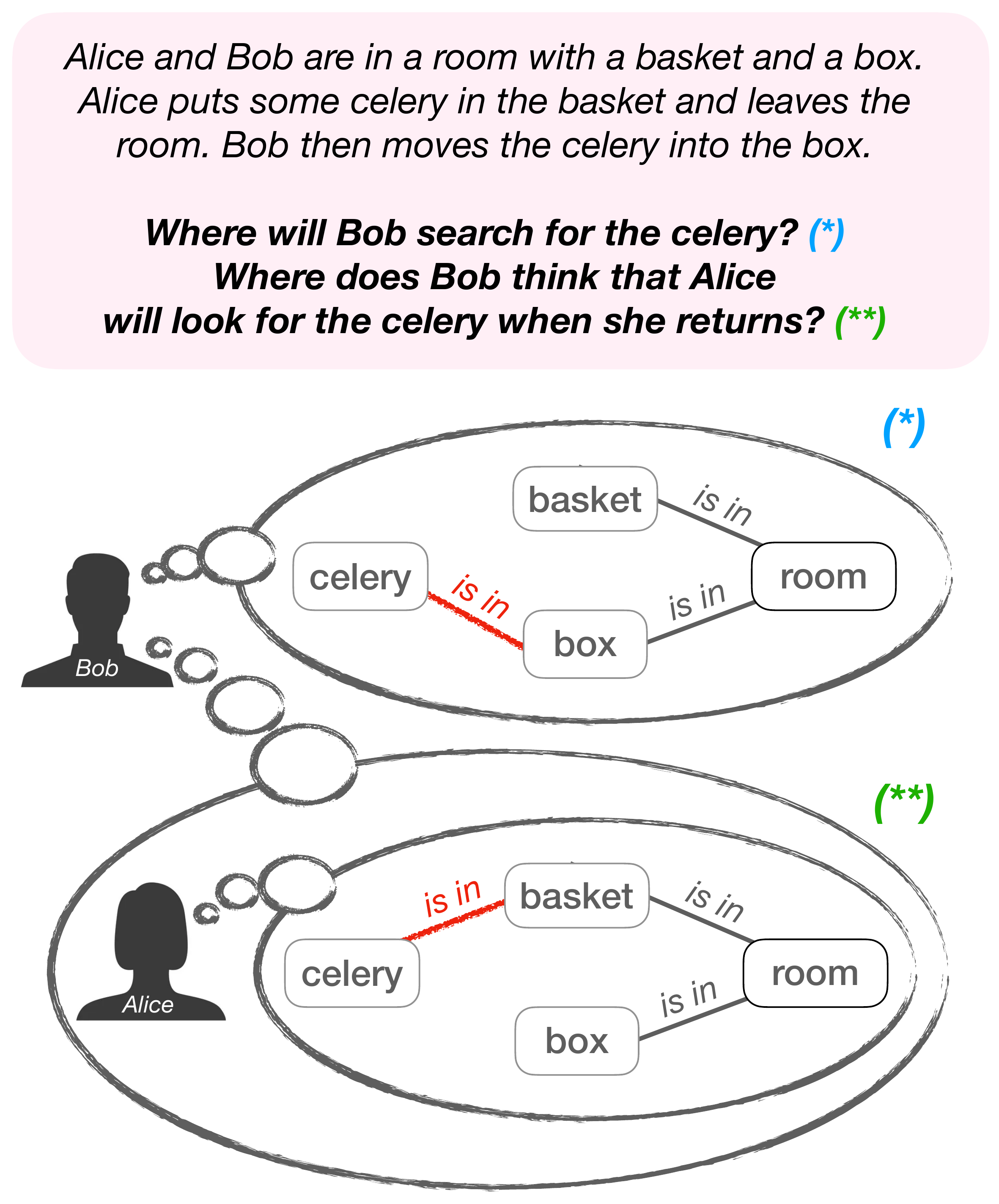}
    \caption{A simple story requiring theory of mind. Note that Alice's belief of the celery's location differs from reality (i.e. Alice holds a \textit{false belief}). Readers must reason that Alice will look for the celery where she left it, and that Bob will make that same assumption. Questions shown require different depths of mental state modeling.
    }
    \vspace{-0.5em}
    \label{fig1:tom_example}
\end{figure}

Reasoning about other people's intentions, desires, thoughts, and beliefs is a cornerstone of human social intelligence. 
Children naturally develop an understanding of every individual's unique mental state and how it might impact their actions~\citep{firth2003developmenttom}. Known as \emph{Theory of Mind (ToM)} \citep{premack1978does}, this ability is crucial for efficient and effective communication. 



Cognitive and literary studies have extensively argued theory of mind's key role in understanding stories, in order to explain and predict each character's actions
~\citep[inter alia]{zunshine2006we, carney2014inference, leverage2010theory, dujin2015narrativetom}. 
As exemplified in Figure~\ref{fig1:tom_example},
%
readers need to model Bob's mental state (called \textit{first-order ToM}), as well as Bob's estimation of Alice's mental state (\textit{second-order ToM}) to answer questions.

Despite recent progress in language understanding abilities,
large language models have been shown to lack theory of mind skills 
\citep{sap2022neural}. Existing efforts to enable them have primarily relied on
supervised methods \citep[e.g.,][]{grant2017HowCM, nematzadeh2018evaluating, arodi-cheung-2021-textual-time}. However, current reading comprehension datasets for theory of mind reasoning are simplistic and lack diversity, leading to brittle downstream models which, as we show, fail in the presence of even slight out-of-distribution perturbations.

We introduce \ourname{}, an inference-time method that improves large language models' theory of mind capabilities by augmenting them with an explicit symbolic graphical representation of each character's beliefs. 
Unlike prior efforts, our approach does not require training and instead divides the problem into simpler subtasks, leveraging off-the-shelf models to solve them, and
carefully consolidating their results. This makes \ourname{} significantly more robust than existing models trained specifically for theory of mind behavior. 


While beliefs about the world state differ among people, most existing work on encoding belief states do not model this behavior relying on singular graphs
~\citep{jansen2022surverytextworlds, jacqmin2022you}. \ourname{}, instead, utilizes 
a \emph{set of graphs}, each representing what the character \textit{$p_1$ thinks that $p_2$ believes that [...] $p_m$ assumes to be the current state of the world}, where $m$ is the maximum reasoning depth as determined by the user. This explicit, recursive mental state representation enables the model to answer questions from the perspective of each character. 
 \ourname{}'s process of selecting and querying a particular character's graph grounds it in cognitive science research arguing theory of mind as an essential mechanism of selective attention~\citep{leslie2004core}. 
Our approach also instills desirable inductive biases, such as object permanence---for example, object locations (represented by edges in the graphs) are assumed to be constant until the method can infer a change. 
Although existing NLP datasets only test up to second-order reasoning (i.e., $m\leq2$), \ourname{} is designed to work at any depth.

\ourname{} dramatically improves the performance of large language models in theory of mind reading comprehension tasks. For example, GPT-3-Davinci's~\citep{brown2020language} accuracy on the ToMi benchmark~\citep{le2019revisiting} increases by 38 absolute points using \ourname{} (yielding 92\% accuracy 
averaging across question types). 
Furthermore, we extend the ToMi test sets with diverse story structures and sentence paraphrases and demonstrate that our approach is significantly more robust than supervised approaches.


\section{Motivation and Background}\label{sec:motivation}

Although large-scale language models have recently shown improvements in some classic theory of mind examples, they are still far from reliably showing theory of mind capabilities~\citep{sap2022neural,yu2022alert, ullman2023large,shapira2023clever}.
%
While the training data for these models includes human-written stories which require theory of mind reasoning, this information is largely implicit and hence difficult for models to learn. ChatGPT and GPT3-Davinci's incorrect answers to Figure~\ref{fig1:tom_example}'s question \#2 are shown below.\footnote{Queried on May 22, 2023 with \texttt{top\_p=1} and \texttt{temperature=0}. Given the non-deterministic and continuously changing nature of these models, exact examples may not produce the same response we report.}



\vspace{0.3em}
\begin{center}
{
\small
\centering
\noindent\fbox{%
    \parbox{0.95\linewidth}{%
        \textbf{ChatGPT \texttt{(gpt-3.5-turbo)}:} Based on the information provided, Bob would likely think that Alice will look for the celery in the box when she returns. Since Bob moved the celery from the basket to the box, he would assume that Alice would expect to find it in its new location. 

\textbf{GPT3 \texttt{(text-davinci-003)}:} Bob will likely think that Alice will look for the celery in the box, since that is where he moved it. 
    }%
}
}
\end{center}
\vspace{0.3em}


Natural stories which make theory of mind explicit are scarce, necessitating automatically generated, template-based datasets like ToM-bAbI \citep{nematzadeh2018evaluating} and ToMi \citep{le2019revisiting}. 
However, templated narratives cover limited types of interactions, and include only simplistic discourse and sentence structures. 
On the other hand, relying on human-generated data, e.g., in situated dialogue~\cite{bara2021mindcraft}, leads to barriers in dataset size due to high annotation costs. 
Moreover, another source of data---text-based games with multiple characters---also faces limitations; in particular, modeling mental states is required mainly to infer intents \citep{zhou2022ai} and to maintain a consistent style of each character \citep{qiu-etal-2022-towards}. 
Rather, in this work, we aim to study and evaluate differences in knowledge and beliefs among multiple characters, traditional \textit{cognitive} aspects of theory of mind.


\begin{figure*}[t!]
\centering
\includegraphics[width=\linewidth]{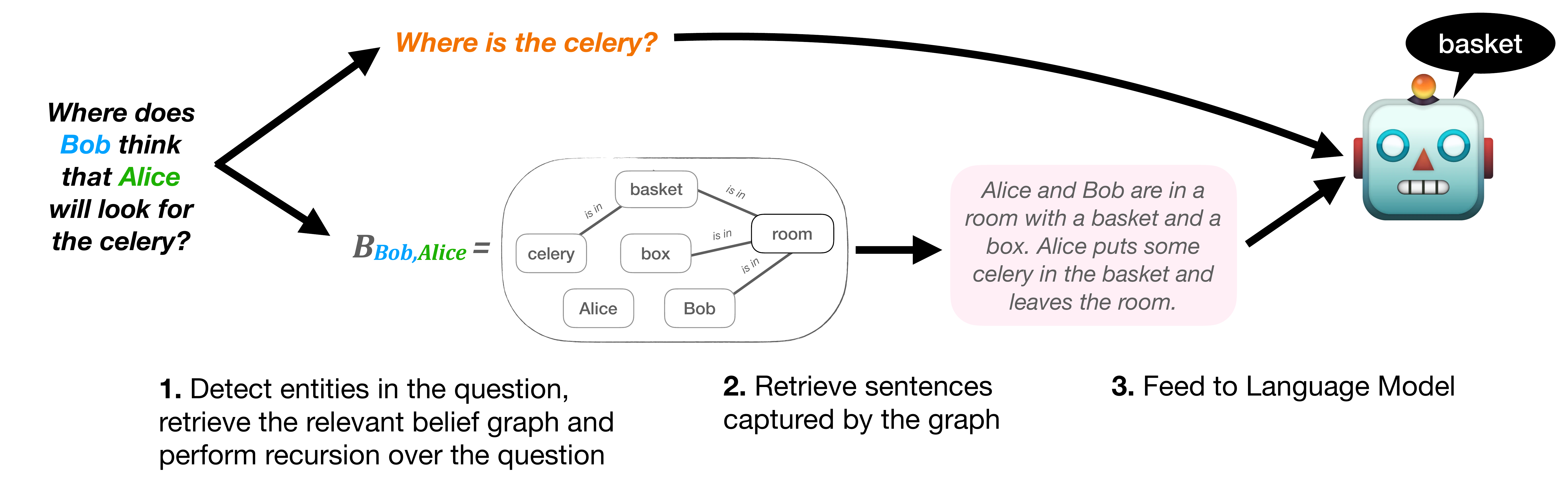}
    \caption{
    Pipeline overview of \ourname{}, a decoding-time algorithm that enhances large language models’ theory of mind capabilities. \ourname{} does not require training: it divides the problem into smaller subtasks and uses off-the-shelf models to solve them. Given a passage, \ourname{} constructs explicit symbolic graphical representations of each character’s belief states (step 1). To answer ToM questions, it retrieves relevant sentences from the graph (step 2) and then queries the LLM in a zero-shot manner (step 3). 
    }
    \label{fig:overview}
\end{figure*}

To the best of our knowledge, the only available datasets for measuring theory of mind in reading comprehension tasks are ToM-bAbI and ToMi. Because of their templated nature, supervised training on them is prone to overfitting to spurious artifacts in the data. While ToMi was developed to counter this behavior in ToM-bAbI by introducing noise in the form of flexible sentence ordering and distractor sentences and characters, we show it still faces the same pitfalls.

Due to theory of mind's inherently implicit nature and limited naturally available data, in this work, we argue against supervision as a way forward and instead call for unsupervised, or inference-time approaches that combine modern neural models and traditional symbolic algorithms.



\section{Methods}\label{sec:methods}


\subsection{\ourname{}: Algorithm Overview}\label{sec:overview}


Our goal is to automatically answer reading comprehension questions given a story involving multiple characters, without requiring any supervised training or fine-tuning on this task. We first introduce key notation, then provide a high-level overview of \ourname{} (Algorithm~\ref{alg:overview}). 

\paragraph{Notation}
We use the term \textit{$k$-th order theory of mind} to refer to an estimate of what a character \textit{$p_1$ thinks that $p_2$ thinks that [...] $p_k$ thinks about the world state}. 
We denote this belief by $B_{p_1,\ldots,p_k}$. We let $k \leq m$, where $m$ is a maximum reasoning depth. This is a user-specified limit, denoting the maximum recursion that the reader is assumed to be capable of performing.
For instance, in Figure \ref{fig1:tom_example}, questions \#1 and \#2 measure 1st- and 2nd-order theory of mind respectively; $B_{\text{Bob}}$ refers to Bob's beliefs about the current world state, and $B_{\text{Bob,Alice}}$ represents Bob's estimation of Alice's beliefs about the world state.
In this work, $B_{p_1,\ldots, p_k}$ only represents beliefs about the current world state, without additional modeling of other characters' mental states, such as their opinions. 

A benefit of this notation is that any belief state can be represented as an $m$-th order one.
We assume that \textit{what $p_k$ thinks that $p_k$ thinks} is equivalent to \textit{what $p_k$ thinks}, and by induction, 
$B_{p_1\ldots p_k}\!\equiv\!B_{p_1,\ldots, p_k, p_k, \ldots, p_k}$, where the last $p_k$ is repeated $m-k$ times. 
We adopt this notation going forward, denoting all states as $m$-th order.
As a conceptual note, the set of belief states $\{B_{p_1\ldots p_k, q_{k+1}\ldots q_{m}} \ | \ \forall q_{k+1},\ldots, q_m \}$ 
represents the mental state from the perspective of $p_1,\ldots,p_k$, using $m-k$ order of theory of mind.

\paragraph{Local and Global Context}
We represent each $B_{p_1\ldots p_k}$ as a graph (a simplified version is depicted in Figure \ref{fig1:tom_example}) where each node represents an entity (e.g. a character, object, room, container) and each edge connects two nodes with a stated relationship in the story.
We construct the graphs by iterating through a story one sentence at a time, and adding both nodes and edges to the graph (\textsc{BeliefTrackingStructure}; described in \S\ref{sec:nested-tracking} and Algorithm~\ref{alg:high-level}).
Each edge is also paired with the sentence from the story from which it was constructed.
We refer to the set of all belief state graphs as the \textit{local contexts}.
We also maintain a \textit{global context} graph, denoted by $G$, which contains the true 
world state. 
$G$ has an identical structure to $B_{p_1\ldots p_k}$. See \ref{app:global_context} for a detailed definition of $G$.

\paragraph{Question Answering} After parsing a story and constructing the complete set of belief-tracking structures, we can use these structures to answer questions by querying the appropriate graph and considering it as the real-world state. 
For example, if the question is ``Where will Bob think that Alice will look for the celery?'',
we retrieve $B_{\text{Bob, Alice}}$, but if instead the question were ``Where will Bob look for the celery?'', we would retrieve $B_\text{Bob}$. 
In both cases, we would ask ``Where is the celery?'' on the retrieved graph. Figure \ref{fig:overview} shows an example of the full pipeline.

Given a question, we identify the relevant characters $p_1, \ldots, p_k$ mentioned in order heuristically, and rephrase the question to ask directly about the world state (\textsc{processQuestion}; owing to the questions' templatic nature in our evaluation data, this approach rephrases all questions correctly
).\footnote{Our explorations show that GPT3 is also capable of rephrasing 
the questions zero-shot (see \S\ref{sec:process-question}), but we refrained from this solution due to budget concerns.}
We then retrieve the corresponding graph; i.e., $B_{p_1, \ldots, p_k}$, of which we can simply ask the question 
``Where is the celery?''. 
To obtain the answer, we first reconstruct a subset $S'$ of sentences in the original story, consisting of those 
represented by the retrieved graph (\textsc{sentencesRepresentedByGraph}).
We then use a large language model $\mathcal{L}$ to answer the simplified question zero-shot given $S'$, using as input the sentences in $S'$ in the same order as they appeared in the original text, and preserving phrasing.\,We optionally further filter $S'$ based on the entities mentioned in the question (\textsc{filterBasedOnQuestion}). An ablation study showed this last step can often be skipped (see Appendix \ref{sec:ablation}). 

\begin{algorithm}[h]
\caption{\ourname{}}\label{alg:overview}
\footnotesize
\begin{algorithmic}
\State $B \gets \textsc{beliefTrackingStructure}(sentences)$
\State $p_1,\!\ldots\!,p_k, question'\!\gets\!\textsc{processQuestion}(question)$
\State $S'\!\gets\!\textsc{sentencesRepresentedByGraph}(B_{p_1,\ldots,p_k})$
\State $S'' \gets \textsc{filterBasedOnQuestion}(S', question)$
\State \Return $S'', question'$ 
\end{algorithmic}
\end{algorithm}
\vspace{-2mm}

\subsection{Computing the Belief Graphs $B_{p_1\ldots p_k}$}\label{sec:nested-tracking}

Assuming each story is told chronologically, \ourname{} processes each sentence $s$ sequentially in two stages (Algorithm~\ref{alg:high-level}). 
First, it extracts all actions in $s$ and updates the global context $G$ from an omniscient point of view while identifying the characters ($\mathcal{W}$) who witnessed actions and world state changes described in the sentence. 
Second, for each witness $w \in \mathcal{W}$, it propagates this new information to update $w$'s local contexts; 
i.e., we only update $B_{p_1, \ldots, p_m}$ with, for $1 \leq i \leq m$, each $p_i \in \mathcal{W}$, and leave the rest unchanged.

\begin{figure}[t]
    \centering
    \includegraphics[width=\linewidth]{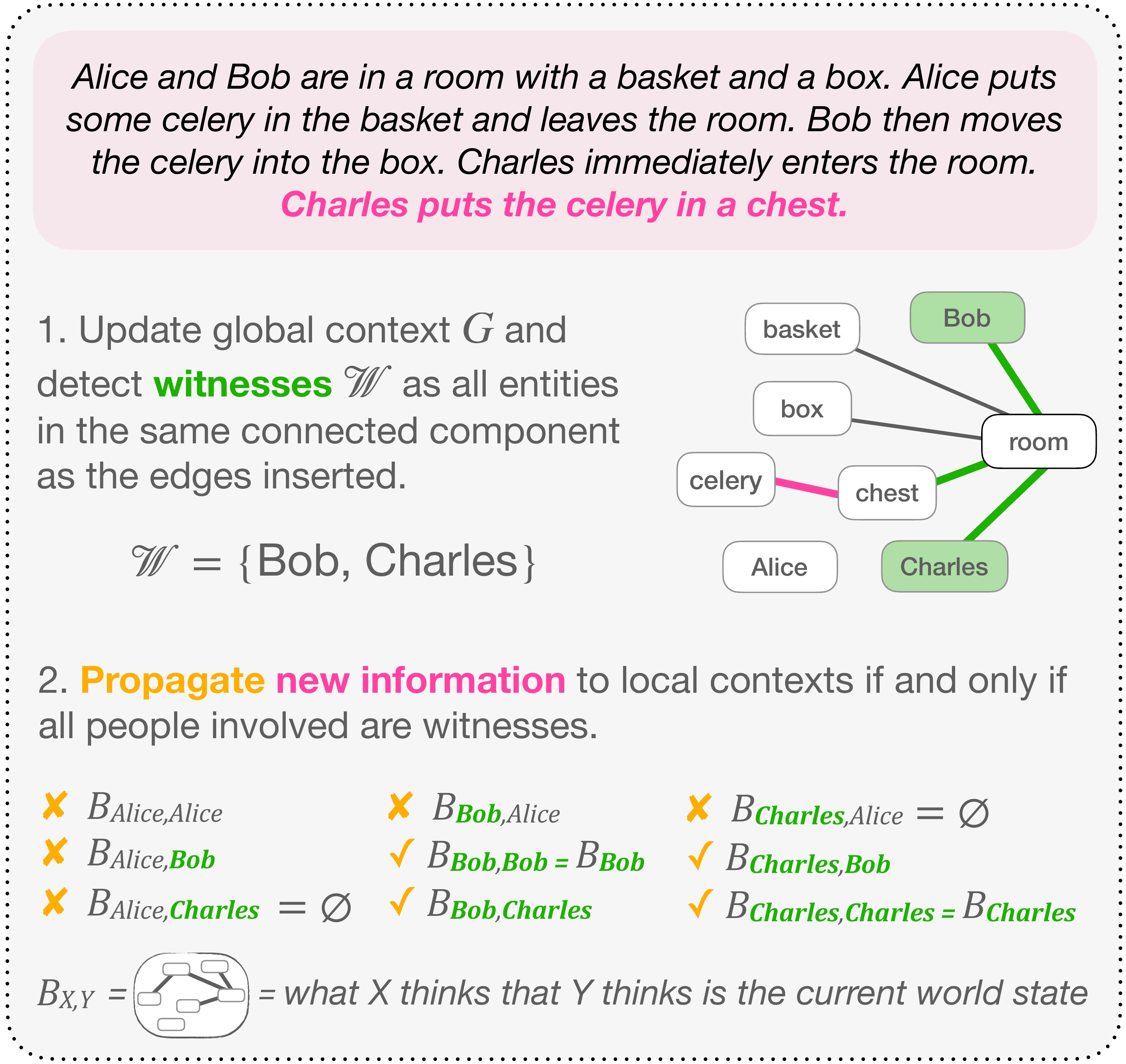}
    \caption{High-level depiction of the belief update procedure for $m=2$. $B_{p_1,\ldots,p_k}$ denotes a graph, and the graph updating procedure is detailed in the main text.}\vspace{3mm}
    \label{fig:witness_propagation}
\end{figure}

As an example, when processing the last sentence in Figure~\ref{fig:witness_propagation}, we update Bob and Charles's state ($B_\text{Bob}$ and $B_{\text{Charles}}$) and the perception of  others' respective state ($B_{\text{Bob,Charles}}$, $B_{\text{Charles, Bob}}$), but we need not update Alice's state, or Bob and Charles's perception of Alice's mental state, because she did not witness the actions described.

\begin{algorithm}[t]
\caption{Belief Tracking 
}\label{alg:high-level}
\footnotesize
\begin{algorithmic}
\Function{beliefTrackingStructure}{$sentences$}
\For{$s \in sentences$}
\State $G, \mathcal{W} \gets \textsc{globalContextUpdate}(G, s)$

\For{ \textbf{all }$[p_1,\ldots,p_m] \in \mathcal{W}^m$}
\State $B_{p_1\!\ldots p_m}\!\gets\!\textsc{localContextUpdate}(B_{p_1\!\ldots p_m},\!G,\!s)$
\EndFor
\EndFor
\EndFunction
\end{algorithmic}
\end{algorithm}


\vspace{3mm}
\subsubsection{Detecting Witnesses, Updating Graphs, and Propagating Knowledge}\label{sec:updating-graphs}


Starting with an empty graph, for each new sentence $s$, we update the global context $G$ by combining off-the-shelf models in four steps (Algorithm \ref{alg:graph-update}; $\textsc{globalContextUpdate}$). 
\begin{inparadesc}
    \item[First,] we detect the existing edges $E$ in $G$ that contradict $s$. This is implemented as detecting Natural Language Inference (NLI) contradictions,
    considering $s$ as the premise, and every edge in $G$ as a hypothesis. 
    \item[Second,] we augment $G$ with new edges and nodes, by first deriving a natural language representation $r$ of the state resulting from the actions described in $s$, and then extract new nodes and edges from $r$ as OpenIE triples~\citep{stanovsky2018supervised}. 
    For example, for ``Bob then moves the celery to the box'', the resulting state $r$ would be the sentence ``The celery is in the box''. 
    To obtain $r$ from $s$, we prompt a language model such as GPT3  (see Appendix~\ref{appendix:resulting-state} for details). 
    After obtaining $r$, we use the corresponding triple (e.g., \texttt{(celery, box, is in)}) to add new nodes and edges to $G$ if not already present (e.g., the nodes ``celery'' and ``box'', and a directed edge connecting them labeled by ``is in'').
    Importantly, we only add edges that represent positive relations between nodes; i.e., there will not be an edge representing ``The celery is not in the box''. 
    \item[Third,] we detect the witnesses $\mathcal{W}$ of the actions described in $s$. Since each character will be a node in $G$, we identify $\mathcal{W}$ as all the characters that are in the same connected component as the newly added edges. 
    \item[Finally,] we remove all edges $E$ that are no longer valid in $G$ as identified by the NLI contradictions. This step is done last to ensure all witnesses are found before their edges are deleted.
\end{inparadesc}


\begin{algorithm}[h]
\caption{World State Beliefs Graphs Update}\label{alg:graph-update}
\footnotesize
\begin{algorithmic}
\Function{globalContextUpdate}{G, s}
\State $E \gets \textsc{detectContradictingEdges}(G, s)$
\State $G \gets G \cup \textsc{triples}(\textsc{resultingState}(s))$
\State $\mathcal{W} \gets \textsc{findWitnesses}(G)$
\State $G \gets G \ \backslash \ E$
\State \Return $G, \mathcal{W}$
\EndFunction
\State
\Function{localContextUpdate}{C, G, s}
\State $E \gets \textsc{detectContradictingEdges}(G, s)$
\State $C \gets C \cup \textsc{triples}(\textsc{resultingState}(s))$
\State $C \gets \textsc{propagateKnowledge}(G, C, s)$
\State $C \gets C \ \backslash \ E$
\State \Return $C$
\EndFunction
\end{algorithmic}
\end{algorithm}



The local contexts ($B_{p_1,\ldots, p_k}$\!) are updated similarly (\textsc{localContextUpdate} in Algorithm~\ref{alg:graph-update}),
except for an additional step of knowledge propagation. While performing an action, a character may implicitly gain information not described in the text. 
For example, when entering a room, a character may gain knowledge of the people and visible objects in the room.
This knowledge (already present in $G$, which tracks the omniscient world state) needs to be propagated to each $B_{p_1,\ldots, p_k}$ with each $p_i\!\in\!\mathcal{W}$.
As $G$ represents the true 
world state, we simplify the problem: 
if a character $p_i$ is in a specific connected component $D$ of $G$, then it possesses all knowledge encoded in $D$.
To model implicit knowledge gain, we add all edges in $D$ to $B_{p_1,\ldots, p_k}$. 
As $D$ represents the latest global context information, we remove from the local context edges that are in $B_{p_1,\ldots, p_k}$ but not in $D$ (representing outdated beliefs about the world state).

\subsection{Notes on Memory Efficiency}
Memory requirements grow exponentially with $m$, the maximum order of theory of mind considered. 
However, $m$ in practice is small, as humans find tasks increasingly challenging as $m$ increases. 
For example, psychological tests for $m=3$ are aimed at teenagers and adults \citep{valle2015theory}. 
All experiments in this work are done with $m=2$, the maximum order of theory of mind reasoning that current datasets evaluate. 
If memory were a concern, one could process the questions first for memory efficiency, and compute only the graphs $B_{p_1,\ldots, p_k}$ required for target queries. 



\section{Fundamental Issues in Existing ToM Datasets}

\paragraph{Construction of ToMi} As introduced in \S\ref{sec:motivation}, 
the sole large-scale theory of mind dataset for reading comprehension tasks is ToMi \citep{le2019revisiting}. 
Barring its added distractor characters and sentences, ToMi strictly mimics the Sally-Anne test, a widely adopted evaluation for assessing children's social cognitive ability to reason about others' mental states \citep{wimmer1983beliefs, baron1985does}. 
Stories are structured are as follows: characters $A$ and $B$ are in a room, and $A$ moves an object from an opaque container to another; $B$ may or may not leave the room before $A$ moves the object. 
$B$ will know the object's new location if and only if they were in the room at the time it was moved. 
Four types of ToM questions are posed: first-order or second-order, probing a character about either a true or a false belief (i.e, belief that matches reality or not). 
ToMi also includes questions probing about reality~\citep[or \textit{zeroth-order} ToM,  ][]{sclar22symmetric} and memory.

ToMi has six types of sentences (i.e. six \textit{primitives}) with set phrasing. These include someone (a) entering or (b) exiting a room; the location of (c) an object or (d) a person; (e) someone moving an object; and (f) someone's opinion about an object (distractors). 
Primitives are combined into stories with a finite list of possible orderings. Despite the limited types of primitives, correctly answering questions requires high-order levels of reasoning.

Templated stories are filled with randomly sampled objects, locations, containers, and rooms from a set list. ToMi implicitly assumes that questions about the story do not depend on these decisions, only on the underlying story template. Yet, in a small-scale human study, we find physical commonsense leads human answers to change, and disagree with ToMi's labels depending on the noun. Table \ref{table:tomi-two-examples} presents an example where the object and container have a large effect on human responses.\footnote{Using Amazon Mechanical Turk, we present 20 humans with the template in Table 1, using either \emph{(hat,box)} or \emph{(apple, pantry)}. Workers are paid \$1 per HIT.}

\begin{table}[t]
\centering
\begin{tabular}{l}
\toprule
\begin{tabular}[l]{@{}l@{}}
1. Oliver entered the front yard. \\
2. Ethan entered the front yard. \\ 
3. Liam entered the kitchen. \\
4. \textbf{objectA} is in the basket. \\ 
5. Ethan exited the front yard. \\ 
6. Ethan entered the kitchen. \\ 
7. Oliver moved \textbf{objectA} to
the \textbf{containerX}. \\
8. Where does Ethan think \textbf{objectA} is? \\
\end{tabular} \\  \midrule
\textit{ToMi Gold Label:} basket \\ \bottomrule
\end{tabular}
\caption{Interpretation of ambiguities in ToMi can be affected by commonsense. 
In the above template, the correct label is that Ethan thinks \textbf{objectA} is in the \textit{basket}, as this is where he last saw it. 
Setting \textbf{objectA} to \emph{hat} and \textbf{containerX} to \emph{box} results in 80\% human accuracy. 
However, setting these to \emph{apple} and \emph{pantry}, accuracy drops to 20\%. 
Physical commonsense suggests the pantry is likely in the kitchen, changing the answer to \emph{pantry}, but regardless of the identity of \textbf{objectA} or \textbf{containerX}, the correct label in ToMi is \emph{basket}. }\label{table:tomi-two-examples}\vspace{-3mm}
\end{table}
%

\paragraph{Resolving Unintentional Ambiguities} ToMi's story construction process often leaves object locations ambiguous, which forces humans to (incorrectly) rely on their physical commonsense. 
For example, the location of the \textit{basket} in line 4 of Table~\ref{table:tomi-two-examples} is ambiguous.
This ambiguity is at times resolved at a later step in the story \citep{arodi-cheung-2021-textual-time}, but it is not true for all cases, and these resolutions were not expressly intended by ToMi's original design.
This complicates the task beyond theory of mind.
For example, in Table \ref{table:tomi-two-examples}, the reader must conclude from \textit{``Oliver is in front yard''}, \textit{``Oliver moved the objectA (...)''}, and \textit{``The objectA is in basket''} that the basket is in the front yard, and hence that Ethan saw it there. 
This requires 3-hop reasoning, and knowing ahead of time that, in ToMi, characters do not change rooms unless explicitly stated.

To solve these unintentional ambiguities and additional 3-hop reasoning requirements, and instead solely measure theory of mind reasoning skills,
we
automatically add a sentence that disambiguates the location of each container immediately after each primitive (c) or (e) 
(e.g., adding \textit{``The basket is in the front yard''} as line 5 in Table \ref{table:tomi-two-examples}). 
Finally, as reported in \citet{arodi-cheung-2021-textual-time,sap2022neural}, ToMi contains some mislabeled second-order questions, which we also correct. 

\section{Experiments}
We experiment with several base LMs, and evaluate each of them both out-of-the-box via zero-shot prompting, and by applying \ourname{} to ToMi stories to produce answers. 
We evaluate Macaw-3B \citep{tafjord2021general}, GPT3-\{Curie,Davinci\} \citep{brown2020language}, Flan-T5-\{XL,XXL\} \citep{chung2022scaling}, LLaMA-\{7B, 13B\} \citep{touvron2023llama}, GPT3.5 \citep{team2022chatgpt}, and GPT4 \citep{openai2023gpt4}. 
We use WANLI \citep{liu2022wanli} for identifying NLI contradictions, and the AllenNLP library \citep{gardner-etal-2018-allennlp} for OpenIE
. We additionally refine each subject and object in extracted triples to remove any stopwords that may be accidentally included by OpenIE. 

We first evaluate \ourname{}'s performance as a plug-and-play method for different base LMs on ToMi (\S\ref{sec:regular-tomi}). We then test whether performance gains are robust to ToMi story structure modifications (\S\ref{sec:robustness}). Finally, we explore \ourname{}'s robustness to linguistic diversity (\S\ref{sec:linguistic-diversity}).

\paragraph{Supervised Models} 
For comparison, we train two supervised models: Textual Time Travel (TTT) \citep{arodi-cheung-2021-textual-time}, and a fine-tuned GPT3-Curie.
TTT is a modification of EntNet \citep{henaff2017tracking} designed for theory of mind tasks; GPT3-Curie is finetuned on 6000 ToMi examples for one epoch.
GPT3-Curie achieves near-perfect performance when finetuned on ToMi (98.5\% accuracy when averaging all questions; Table \ref{table:perf-regular-test-set}). 
Interestingly, GPT3-Curie achieves a higher accuracy than the theory of mind-motivated TTT (accuracy 92.3\%). We explore model robustness in \S\ref{sec:robustness}.


\subsection{In-Domain Evaluation}\label{sec:regular-tomi}

\begin{figure}[t]
    \centering
    \includegraphics[width=0.96\linewidth]{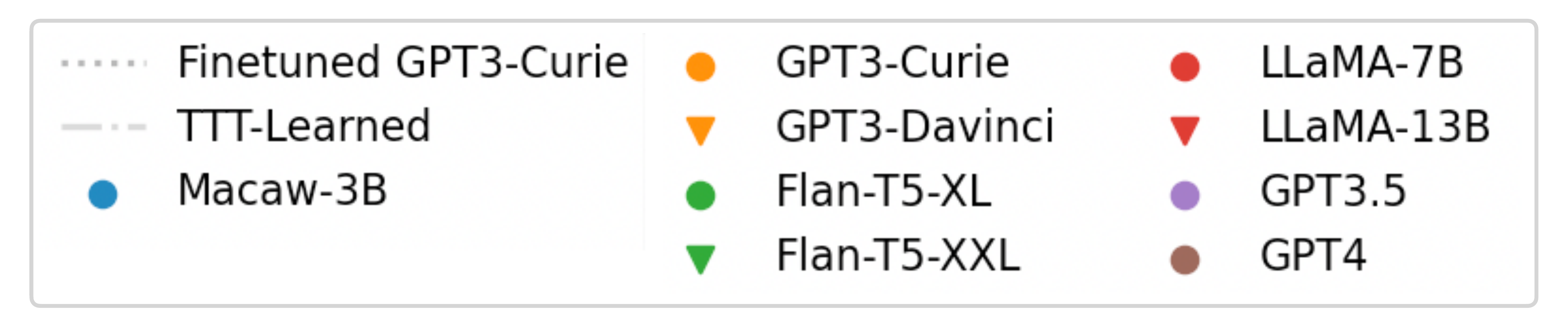}
    \includegraphics[width=\linewidth]{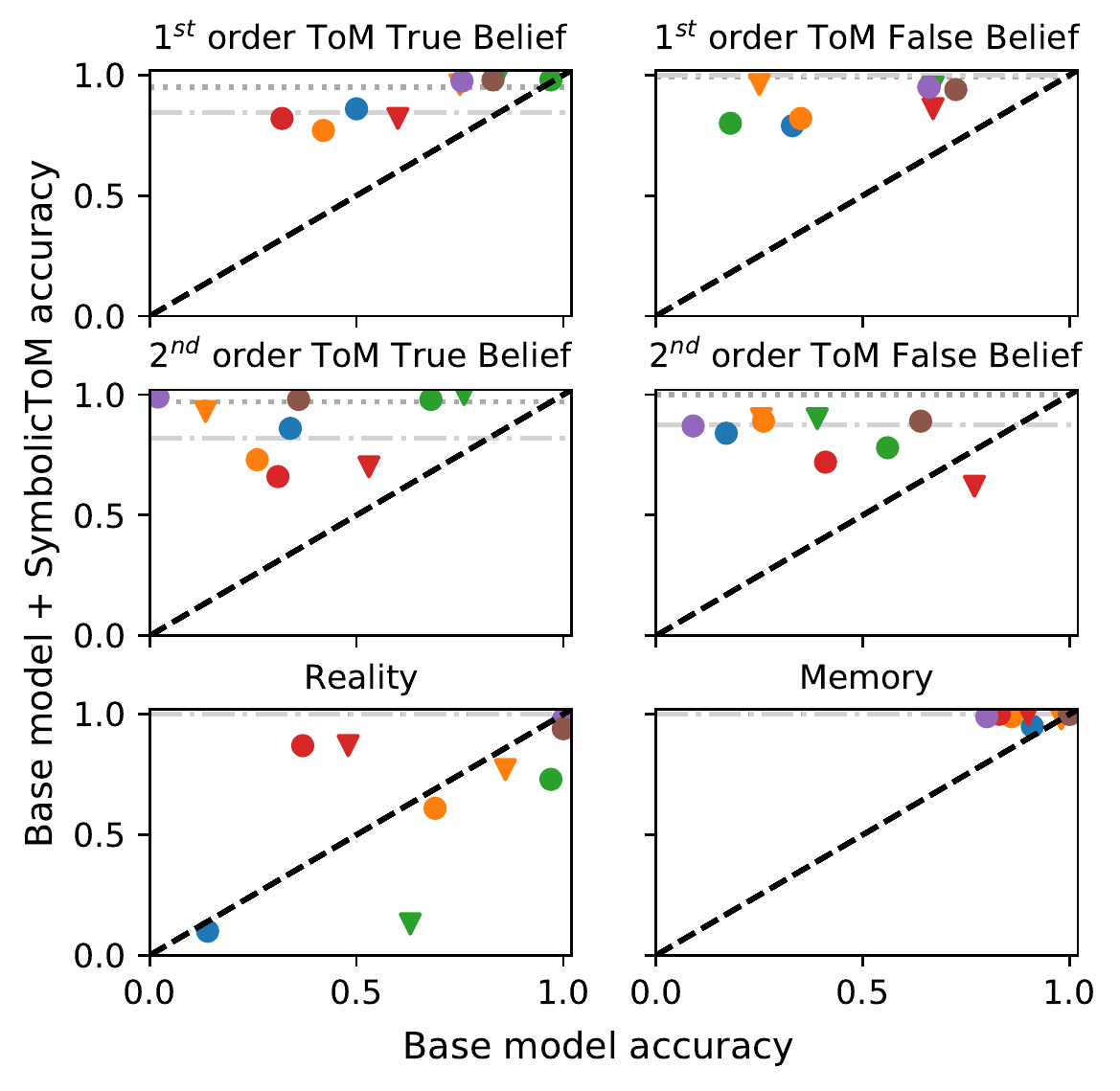}
    \caption{Accuracy for each ToMi question type and base model (higher is better). Dots in the upper triangle have higher performance with \ourname{} than the base model out-of-the-box. Horizontal lines give supervised models' performance. Full results in Table~\ref{table:perf-regular-test-set}.}
    \label{fig:performance-by-question}
\end{figure}

We evaluate all base LMs comparing their performance out-of-the-box, versus when adding \ourname{}. Figure \ref{fig:performance-by-question} shows results by question type, showing dramatic improvements for all theory of mind questions: +62 points in accuracy for first-order false-belief questions for Flan-T5-XL, +78 points in accuracy for second-order false-belief questions for GPT3.5,
among other improvements.
In addition, we observe all models maintain near-perfect performance with and without \ourname{} in memory questions. Supervised models show high accuracy for all question types.

We only see significant decreases in performance for reality questions in Flan-T5 models. This can be partially attributed to the questions' phrasing: questions are posed as ``Where is the celery \textit{really}?''. 
Removing \textit{really} results in 96\% accuracy for Flan-T5-XL. 
Flan-T5-XXL empirically shows a bias towards providing a room rather than container as an answer when only one container is mentioned, which is often the case for \ourname{}-filtered stories. 
Rooms are invalid answers in ToMi. 
An ablation on the final filter function of Algorithm \ref{alg:overview} suggests that keeping more containers in the final story reduces this bias and still yields significant improvements for false-belief questions across all models (see \S\ref{sec:ablation}). Besides \textit{reality} questions, Flan-T5-XXL with \ourname{} achieves results comparable to the supervised TTT.


\subsection{Story Structure Robustness Test Sets}\label{sec:robustness}
We create three test sets by modifying ToMi's stories structures without adding new types of actions or linguistic diversity. These tests were only evaluated once, after finishing development of \ourname{}. Test sets are defined below. See Appendix \ref{appendix:example-test-sets} for concrete examples. 

\begin{table}[t]
\centering
\normalsize
\begin{tabular}{@{}cccc@{}}
\toprule
             & $D_1$            & $D_2$            
             & $D_3$            \\ \midrule
\rowcolor{verylightgray} \multicolumn{4}{l}{\textit{Off-the-shelf models}} \\
Macaw-3B     & \phantom{0}8 & 12 
& 30 \\
Flan-T5-XL     & 86	& 51 
&  68 \\
Flan-T5-XXL     &  69 &  59 
& 52 \\
GPT3-Curie   & 37 & 39 
& 57 \\
GPT3-Davinci & 20 & 25 
& 39 \\ 
GPT3.5\footnote{Low scores are due to the model refusing to answer, e.g. answering \textit{``There is no information in the given text to determine where Bob thinks Alice searches for the celery.''}} & \phantom{0}1 & \phantom{0}0 &  48 \\ 
GPT4 & 58 & 62 & 97 \\ 
LLaMA-7B & 17	& 17 & 17 \\ 
LLaMA-13B & 26 & 36 & 37 \\ 
\midrule
\rowcolor{verylightgray} \multicolumn{4}{l}{\textit{\ourname{} + Off-the-shelf models}} \\
Macaw-3B     & \phantom{0}89 {\small (+81)} & \phantom{0}71 {\small (+60)}  
& \phantom{0}70 {\small (+41)} \\
Flan-T5-XL     & \phantom{0}76 {\small (-10)}	& \phantom{0}96 {\small (+46)} 
& \textbf{100} {\small (+33)} \\
Flan-T5-XXL     &  \phantom{0}93 {\small (+24)}  &  \textbf{100}  {\small (+41)} 
& \textbf{100} {\small (+49)}  \\
GPT3-Curie   & \phantom{0}84 {\small (+48)} & \phantom{0}81 {\small (+42)} 
& \phantom{0}73 {\small (+16)} \\
GPT3-Davinci & \phantom{0}92 {\small (+73)}  & \phantom{0}91 {\small (+66)}  
& \phantom{0}90 {\small (+50)}  \\ GPT3.5 & \textbf{100} \textbf{\small \color{bettergreen} (+99)} &  \textbf{100} \textbf{\small \color{bettergreen} (+99)} &  \phantom{0}99 {\small (+51)} \\ 
GPT4 & \textbf{100} {\small (+42)} &  \textbf{100} {\small (+38)} &  \textbf{100} {\small (\phantom{0}+4)} \\ 
LLaMA-7B & \phantom{0}99 {\small (+82)} & \phantom{0}92 {\small (+75)} &  \phantom{0}88 \textbf{\small \color{bettergreen} (+71)} \\ 
LLaMA-13B & \phantom{0}78 {\small (+52)} & \phantom{0}84  {\small (+48)} & \phantom{0}84 {\small (+47)} \\ 
\midrule
\rowcolor{verylightgray} \multicolumn{4}{l}{\textit{Supervised models}} \\
{\small TTT}  & 49 & 65 
& 78 \\ 
{\small Finetuned GPT3} & 51 & 68 
& 32 \\ \bottomrule
\end{tabular}
\caption{Precision using \ourname{} on all questions from 100 stories for each of the modified test sets $D_i$. Supervised models were trained on ToMi; all others do not require training. Parenthesis reflect differences between using and not using \ourname{}: \textbf{bold} reflects higher overall performance, and {\color{bettergreen} \textbf{green}} reflects the highest net improvements when using \ourname{}. 
}\label{tab:custom-test-sets}
    \vspace{-0.5em}
\end{table}

\paragraph{Double Room False Belief Story ($D_1$)} Two false belief substories involving the same two characters $p_1, p_2$ are concatenated to yield a longer, more complex story. Each substory has different objects being moved, across different containers. The system is probed using all four combinations of second-order theory of mind questions involving the two characters and locations. Questions are evenly split between the first and second substory.
\paragraph{Three Active Characters Story ($D_2$)} Three characters $p_1,p_2,p_3$ are in the same room, where 
an object $o_1$ and three containers $c_1, c_2, c_3$ are available. The story is as follows: $p_2$ leaves before $p_1$ moves $o_1$ from $c_1$ to $c_2$, but $p_3$ witnesses the move. Then, $p_1$ leaves the room. Later, $p_3$ moves the object to container $c_3$ without any witnesses. The system is probed using all combinations of second-order theory of mind questions.
\paragraph{Multiple Object Movements Across Four Containers ($D_3$)} Two characters $p_1,p_2$ are in a room, with a single object, and four containers $c_1,\ldots,c_4$. $p_1$ moves the object from $c_1$ to $c_2$ and right before leaving the room, $p_2$ enters. $p_2$ then moves the object to $c_3$, and then $c_4$. We probe with all first and second-order theory of mind questions.


\paragraph{Results}
Supervised models  significantly overfit to ToMi's original story structures (Table \ref{tab:custom-test-sets}). In contrast, all models had high accuracy when equipped with \ourname{}, especially larger models
, such as GPT3.5, LLaMA-\{7B,13B\}, among\,others.

$D_2$ may also be used to test third-order ToM reasoning, asking questions such as ``Where does $p_1$ think that $p_2$ thinks that $p_1$ will search for the $o_1$?''. Third-order ToM is a reasoning depth currently untested by available NLP benchmarks. \ourname{} consistently enhances the performance of off-the-shelf LLMs and outperforms supervised methods in the third-order ToM setting.  See details in Appendix \ref{app:third_order_tom}. This experiment showcases how extensions of ToMi may be used to test higher-order reasoning. This is the first approach towards testing third-order ToM in LLMs; a benchmark to comprehensively test such order of reasoning exceeds the scope of this paper.

\subsection{Paraphrasing Robustness Evaluation}\label{sec:linguistic-diversity}

\begin{figure}[t]
    \centering

    \includegraphics[width=\linewidth]{images_camera_ready/labels_symbolictom.pdf}
    \includegraphics[width=\linewidth]{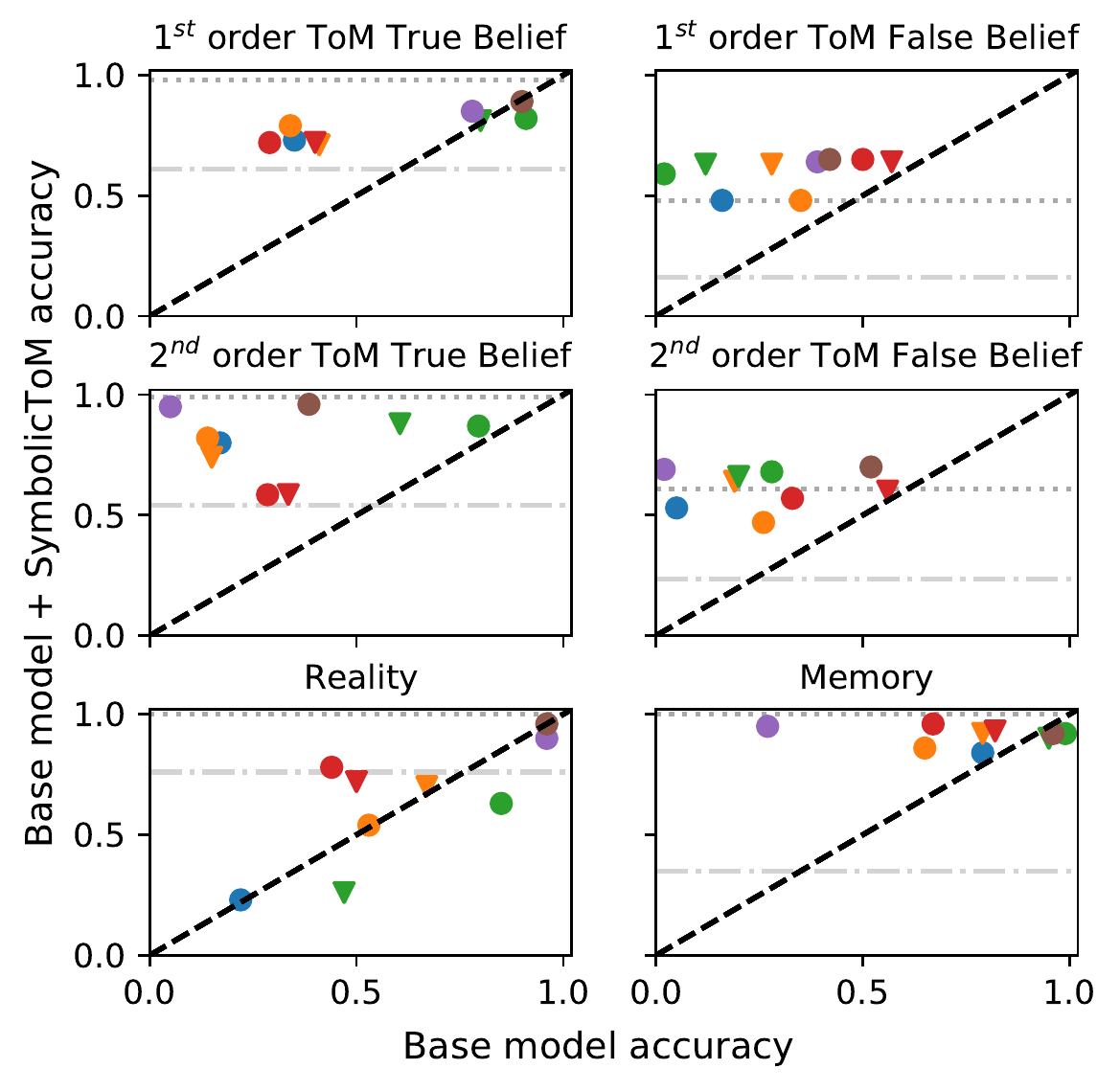}
    \vspace{-1.4em}
    \caption{Results for ParaphrasedToMi when prompting GPT3 as implementation of \textsc{resultingState} (Davinci for all except for Curie). Dots in the upper triangle imply performance with \ourname{} is higher than using the base model out-of-the-box. Horizontal lines reflect supervised models' performance (higher is better).}\vspace{-1em}
    \label{fig:performance-by-question-linguistic-diversity}
\end{figure}

We assess the robustness of all models when utilizing various wordings for each sentence. We reword all templates using GPT3-Davinci, utilizing different choices of objects, rooms, and names, and manually excluded incorrect paraphrases. The resulting dataset---ParaphrasedToMi---exhibits much greater complexity, as these rewordings can express actions in a less  straightforward way. All paraphrases are shown in Appendix \ref{appendix:linguistic-diversity}. 

Figure \ref{fig:performance-by-question-linguistic-diversity} demonstrates significant performance decreases for supervised models transferring to ParaphrasedToMi. TTT's average accuracy drops 54 points from ToMi, with losses across all question types. Finetuned GPT3 exhibits significant losses in false-belief questions (-40 average accuracy) but is robust for other question types.

Methods without supervision also suffer significant losses, but \ourname{} still results in large improvements for theory of mind questions. Models equipped with \ourname{} perform significantly better than the supervised TTT model across all theory of mind questions. ParaphrasedToMi is significantly more difficult for \ourname{} since it triggers more errors in edge removal (due to errors in NLI classification), as well as errors in edge insertion (due to errors in the resulting state's triple extraction). Although computing \textsc{resultingState} by prompting the base LMs was successful with original phrasings (as defined in \S\ref{sec:updating-graphs}), we observed differences in robustness when prompting with paraphrases. We found implementing \textsc{resultingState} with GPT3 reliable, and thus we use it for all models. Results using other models are included in \S\ref{sec:linguistic-diverisity-using-self-models}: false-belief performance is even better for models like LLaMA, GPT3.5, or GPT4. 

\section{Related Work}

\paragraph{Existing Approaches} 
Classical reasoning tasks require achieving some goal, e.g., proving a statement, given a set of facts and universally valid rules
~\citep[e.g.,][]{tafjord-etal-2021-proofwriter}. 
A common approach is to decompose the target reasoning task into subtasks, for example by using off-the-shelf LMs~\cite{creswell2022selection,kazemi2022lambada, nye2021improving}.
We use a similar technique in \ourname{}, breaking the higher-level reasoning task into graph reasoning subtasks. Nonetheless, these approaches cannot be simply ported to our domain: stories' facts (i.e. the world state) change over time and are not universally accessible to all characters, and commonsense rules and assumptions like object permanence must made explicit. 
\ourname{}'s design addresses these challenges by maintaining and updating graphs about facts and beliefs as a story progresses. 

In scenarios where world state changes over time, such as in text-based games, existing approaches maintain and update structured world representations as the world state changes~\cite{ammanabrolu2021learning,adhikari2020learning}.
However, while these approaches could potentially be applied in our scenario to update $G$, they would not address the problems of multiple-belief representation or knowledge propagation to witnesses' graphs, with some approaches even being explicitly impossible for modeling second-order ToM~\cite{qiu-etal-2022-towards}.


\paragraph{ToM beyond NLP} 
Theory of mind is also crucial in multi-agent reinforcement learning \cite{rabinowitz2018machine}, including in bidirectional symbolic-communication \citep{wang2021tom2c, sclar22symmetric}, unidirectional natural-language settings~\citep{zhu2021few}; and recently, by combining reinforcement learning, planning, and language, to create a human-level Diplomacy player \citep{meta2022human}. 
It has also received increased attention in human-computer interaction \citep{wang2021hcitom} and explainable AI \citep{AKULA2022103581}. 

Psychologists divide theory of mind into two types of reasoning: affective (emotions, desires) and cognitive (beliefs, knowledge) \citep{shamay2010role}, with the former developing earlier in children \citep{wellman2014making}. 
Our work focuses on the latter, 
but the principle of multiple belief representation could also be applied to affective theory of mind reasoning.
Existing work has shown that humans are proficient at second-order or higher false-belief reasoning, also referred to as \textit{advanced ToM}~\cite{advancedtom2017}, with evidence that we can perform even third- and fourth-order reasoning~\citep{valle2015theory, osterhaus2016scaling}.
While, to best of our knowledge, no dataset requires beyond second-order ToM,
\ourname{} explicitly models the recursive reasoning that supports queries of any reasoning order. 


\section{Conclusions}

Theory of mind is an essential social intelligence ability. Developing agents with theory of mind is requisite for a wide range of applications, including reading comprehension, tutoring, dialogue, personalization, 
and negotiation. 
For example, in reading comprehension settings (and broadly for natural language understanding), having a multi-level understanding of texts is crucial for providing meaningful and contextualized answers: stories often rely on theory of mind reasoning to create conflict (e.g., in murder mysteries, drama, and romances, as in the final acts of \textit{Romeo and Juliet}).

We present \ourname{}, a plug-and-play method to enable theory of mind reasoning in language models 
via explicit symbolic representations in the form of nested belief states. \ourname{} requires no training or fine-tuning, 
a key aspect for a domain with scarce supervised data and limited success in learning from massive unlabeled text alone. 
With experiments on reading comprehension tasks, our approach demonstrates dramatic improvement in the accuracy of base language models, especially for false-belief scenarios.

We also show that, in contrast to supervised methods, \ourname{} is highly robust to story perturbations and out-of-domain inputs where supervised methods suffer significant degradations~\citep[as in, e.g.,][]{yu2022alert}.
\footnote{As a part of out-of-domain testing, we also create a more challenging version of the available ToM datasets, available at \url{https://github.com/msclar/symbolictom} along with a corrected version of ToMi. 
} 
Our results show the promise of augmenting neural language models with symbolic knowledge
for improving their social reasoning skills. We leave to future work to investigate similar approaches for other types of social intelligence; as well as develop new datasets that cover a more diverse set of interactions.



\section*{Limitations}


\ourname{} assumes stories are written chronologically, which may not hold for some human-written stories. This may be alleviated using time-stamping models like \citet{faghihi2021time}.
Furthermore, since we use off-the-shelf models (WANLI \citep{liu2022wanli} and OpenIE \citep{stanovsky2018supervised}) to create and update the graphs, the presented approach may propagate errors as revealed in the linguistic diversity experiments. However, these issues can be largely alleviated by using more sophisticated models, even the LLMs like GPT3 themselves. We do not experiment with them due to budgetary restrictions.

Currently, all NLP datasets available for theory of mind reasoning describe Sally-Anne tests. In these datasets, the concept of large distances is absent, meaning that anyone specified to be in a location is assumed to be a witness of the actions that occur there. 
This assumption can be violated in realistic settings. For example,
\textit{``Anne is in the USA''} does not imply she is a witness to every action happening in the USA.
In future work, this approach can be improved by refining the witnesses detection algorithm to incorporate physical commonsense reasoning. We could also refine the witness detection algorithm by sampling paths between the inserted edge and each node referring to a person, to query an LM directly on that substory by asking if the person witnessed the action. To be able to test both of these ideas, we would need to obtain new theory of mind datasets with significantly more types of interactions and physical commonsense in the~stories.

\section*{Ethics Statement}

Theory of mind research at its core deals with reasoning about the mental states of others. In this work, we focus on reading comprehension, a task which 
can similarly be exposed to ethical concerns: for example, when a model makes erroneous predictions about the mental states of characters in the description, when it is misused to reason about private situations, and when it makes predictions which reinforce social biases. This issue can be exacerbated if the characters are actual people. In this work, however, we experiment with simple, prototypical character references from a public dataset, and not with actual people. This decision is intentional. Furthermore, we focus on reasoning about physical objects and observers' knowledge about their location in space, which is less prone to ethical concerns. This data can nonetheless lead to biased decisions, such as imbalanced decisions correlated with social attributes like gender (often correlated with names). 
Future work in this area may include scenarios with more realistic human-agent interaction, such as dialogue tasks, where parties involved may not have the same incentive structure. 
These scenarios will need to be handled with special care as they could lead to agents learning to deceive humans by exploiting a predicted (lack of) knowledge
. The state-of-the-art in machine theory of mind is still far from these capabilities, but we believe it is important to consider these risks when designing experiments.

\section*{Acknowledgements}
We thank Lucille Njoo and Tianxing He for the valuable discussions, and Akshatha Arodi for the support in running the Textual Time Travel code base. 
S.K. gratefully acknowledges support from Google Ph.D.~Fellowship.\,We also thank OpenAI for providing academic access to their language model API. This material is based upon work partly funded by the DARPA CMO under Contract No.~HR001120C0124, by DARPA MCS program through NIWC Pacific (N66001-19-2-4031), by NSF DMS-2134012, by NSF CAREER Grant No.~IIS2142739, and an Alfred P.~Sloan Foundation
Fellowship.
Any opinions, findings and conclusions or recommendations expressed in this material are those of the authors and do not necessarily state or reflect those of the United States Government or any agency thereof.



\bibliography{anthology,custom}
\bibliographystyle{acl_natbib}

\appendix

\section{Additional Details on \ourname{}}

\subsection{Detailed Description of Information Contained in Global Context $G$}\label{app:global_context}

In the main paper, we define $G$ as a graph containing the true world state (as opposed to beliefs about the current world state). This means that $G$ will represent where people and objects are truly located, regardless of beliefs. $G$ will in general contain only the \textit{observable} true world state. Thus, information passed verbally would not be stored in the global context (e.g. someone speaking in a room is not observable after they finished talking), and would instead be stored in the local contexts of the people that heard the speech. Since verbal interactions are not tested by available datasets, this distinction is not relevant in ToMi.


\subsection{Prompts for Resulting State Extraction}\label{appendix:resulting-state}

For GPT3-Curie we 2-shot prompt with the following prompt (both for original and linguistic diversity experiments):\\
\noindent\texttt{John quit his job. The resulting state after this action is that John no longer has a job.\textbackslash n\textbackslash nJohn signed a contract. The resulting state after this action is that the contract is signed.\textbackslash n\textbackslash n\textbf{<sentence>}. The resulting state after this action is that}\\

We find that GPT3-Davinci, Flan-T5-XL, GPT3.5, and GPT4 are able to zero-shot answer to this subtask just by describing the instruction, but smaller models benefit from few-shot. We were unable to query Macaw for this task, so we instead rely on GPT3-Curie, a model of comparable size. Zero-shot instruction is as follows:

\texttt{\textbf{<sentence>}. What is the resulting state after this action? Do not add new information. The resulting state after this action is that}

We observe that GPT3 is significantly more robust to paraphrases than Flan-T5: Flan-T5 models are poor at detecting the resulting state for florid paraphrases, although the original phrasings are a straightforward task for Flan-T5.

Larger models like GPT3.5 and GPT4 are able to perform the task well zero-shot, similarly to GPT3; LLaMA models require fewer demonstrations than Flan-T5. We ran all main experiments implementing Resulting State Extraction with GPT3.

\subsection{Solving \textsc{processQuestion} using GPT3}\label{sec:process-question}

\begin{figure}[h!]
    \centering
    \includegraphics[width=\linewidth]{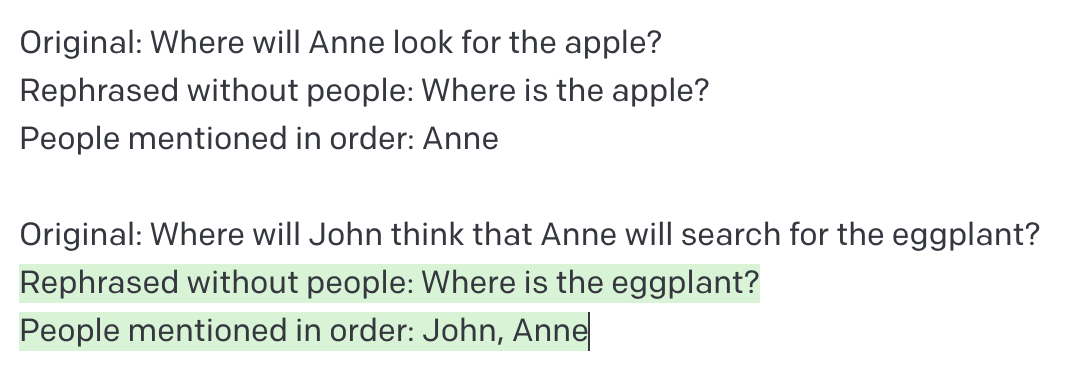}
    \caption{GPT3 shows one-shot generalization abilities from first-order to second-order questions.}
    \label{fig:screenshot-rephrase-question}
\end{figure}

Our explorations suggest that GPT3 (Curie and GPT3-Davinci \texttt{text-davinci-002}---the version used in all our experiments) can successfully extract entities and rephrase the question. See Figure \ref{fig:screenshot-rephrase-question} for an example prompt.

\section{Details on Out-Of-Domain Evaluation}

\subsection{Linguistic Diversity Per ToMi Template}\label{appendix:linguistic-diversity}
\begin{table}[h]
\center
\begin{tabular}{@{}cc@{}}
\toprule
\textit{Sentence type}  & \textit{Count} \\ \midrule
Object's Position        & 38             \\
Distractor Negative Sentiment   & 36             \\
Distractor Positive Sentiment   & 31             \\
Person Entered Room                 & 21             \\
Person Exited Room                  & 19             \\
Person Moved Object           & 18             \\
Person's Position        & 9              \\ \bottomrule
\end{tabular}
\caption{Number of paraphrases per original sentence template. Paraphrases were obtained from prompting GPT3-Davinci (\texttt{text-davinci-002}).}
\end{table}

\subsubsection{All Paraphrases of \texttt{PersonX entered the RoomY}.}
\noindent\texttt{PersonX entered the RoomY.\\
PersonX approached the RoomY.\\
PersonX arrived at the RoomY.\\
PersonX arrived in the RoomY.\\
PersonX bounded into the RoomY.\\
PersonX came by the RoomY.\\
PersonX came into the RoomY.\\
PersonX came to the RoomY.\\
PersonX crept into the RoomY.\\
PersonX entered the RoomY.\\
PersonX leapt into the RoomY.\\
PersonX showed up at the RoomY.\\
PersonX shuffled into the RoomY.\\
PersonX sidled into the RoomY.\\
PersonX slithered into the RoomY.\\
PersonX stepped into the RoomY.\\
PersonX tiptoed into the RoomY.\\
PersonX visited the RoomY.\\
PersonX walked into the RoomY.\\
PersonX went into the RoomY.\\
PersonX went to the RoomY.
}

\subsubsection{All Paraphrases of \texttt{PersonX exited the RoomY}.}

Prompted with the prompt: \textit{Find 30 alternative ways of expressing the following sentence: Abigail exited the bedroom.} and manually filtering results (with this and other name/location selection.

\noindent\texttt{PersonX exited the RoomY.\\
PersonX left the RoomY.\\
PersonX walked out of the RoomY.\\
PersonX stepped out of the RoomY.\\
PersonX departed the RoomY.\\
PersonX went out of the RoomY.\\
PersonX came out of the RoomY.\\
PersonX emerged from the RoomY.\\
PersonX quit the RoomY.\\
PersonX took off from the RoomY.\\
PersonX bolted from the RoomY.\\
PersonX flew from the RoomY.\\
PersonX ran from the RoomY.\\
PersonX sprinted from the RoomY.\\
PersonX jogged from the RoomY.\\
PersonX hurried from the RoomY.\\
PersonX crawled from the RoomY.\\
PersonX crept from the RoomY.\\
PersonX tiptoed from the RoomY.}

\subsubsection{All Paraphrases of \texttt{The Object1 is in the Container1.}}

Prompted with \texttt{Object1=apple}, \texttt{Container1=\{fridge, envelope, bathtub\}}. Then filtered to remove object-specific wording.\\

\noindent\texttt{The Object1 is in the Container1.\\
The Object1 is stored in the Container1.\\
The Object1 is kept in the Container1.\\
The Object1 is located in the Container1.\\
The Object1 is situated in the Container1.\\
The Object1 is set in the Container1.\\
The Object1 is placed in the Container1.\\
The Object1 is found in the Container1.\\
The Object1 is positioned in the Container1.\\
The Object1 is set upon in the Container1.\\
The Object1 is put in the Container1.\\
The Object1 is laid in the Container1.\\
The Object1 is deposited in the Container1.\\
The Object1 is stationed in the Container1.\\
The Object1 is put to rest in the Container1.\\
The Object1 is set to rest in the Container1.\\
The Object1 is rested in the Container1.\\
The Object1 is set aside in the Container1.\\
The Object1 is stowed in the Container1.\\
The Container1 contains the Object1.\\
The Object1 is inside the Container1.\\
The Object1 is within the Container1.\\
The Container1 is where the Object1 is.\\
The Container1 has the Object1.\\
The Container1 is holding the Object1.\\
The Container1 is keeping the Object1.\\
The Container1 is safeguarding the Object1.\\
The Container1 is storing the Object1.\\
The Container1 has the Object1 within it.\\
The Container1 has the Object1 inside of it.\\
The Container1 is holding the Object1 within it.\\
The Container1 is keeping the Object1 inside of it.\\
The Container1 is safeguarding the Object1 inside of it.\\
The Container1 is storing the Object1 inside of it.\\
There is a Object1 in the Container1.\\
A Object1 is in the Container1.\\
The Container1 has a Object1 in it.\\
Inside the Container1 is a Object1.
}

\subsubsection{All Paraphrases of \texttt{PersonX moved the Object1 to the Container1.}}

\noindent\texttt{PersonX moved the Object1 to the Container1.\\
PersonX relocated the Object1 to the Container1.\\
PersonX transferred the Object1 to the Container1.\\
PersonX shifted the Object1 to the Container1.\\
PersonX placed the Object1 in the Container1.\\
PersonX set the Object1 in the Container1.\\
PersonX put the Object1 in the Container1.\\
PersonX stowed the Object1 in the Container1.\\
PersonX stored the Object1 in the Container1.\\
PersonX hid the Object1 in the Container1.\\
PersonX shoved the Object1 into the Container1.\\
PersonX pushed the Object1 to the Container1.\\
PersonX carried the Object1 to the Container1.\\
PersonX conveyed the Object1 to the Container1.\\
PersonX led the Object1 to the Container1.\\
PersonX transported the Object1 to the Container1.\\
PersonX brought the Object1 to the Container1.\\
PersonX took the Object1 to the Container1.}

\subsubsection{All Paraphrases of \texttt{PersonX is in the RoomY.}}

\noindent\texttt{PersonX is in the RoomY.\\
PersonX is inside the RoomY.\\
PersonX is located in the RoomY.\\
PersonX is situated in the RoomY.\\
PersonX is present in the RoomY.\\
PersonX is to be found in the RoomY.\\
PersonX is contained in the RoomY.\\
The RoomY holds PersonX.\\
The RoomY shelters PersonX.}

\subsubsection{All Paraphrases of positive distractor sentences}
\noindent\texttt{PersonX has a bad case of Object1 fever.\\
PersonX is Object1 crazy.\\
PersonX is Object1-crazed.\\
PersonX is Object1-obsessed.\\
PersonX is a Object1 fiend.\\
PersonX is a Object1 maniac.\\
PersonX is a Object1-aholic.\\
PersonX is always thirsty for a Object1.\\
PersonX is besotted with the Object1.\\
PersonX is captivated by the Object1.\\
PersonX is charmed by the Object1.\\
PersonX is crazy about the Object1.\\
PersonX is crazy for the Object1.\\
PersonX is eager for the Object1.\\
PersonX is enamored with the Object1.\\
PersonX is enthusiastic about the Object1.\\
PersonX is entranced by the Object1.\\
PersonX is fascinated by the Object1.\\
PersonX is fond of the Object1.\\
PersonX is in love with the Object1.\\
PersonX is infatuated with the Object1.\\
PersonX is keen on the Object1.\\
PersonX is mad about the Object1.\\
PersonX is never seen without a Object1.\\
PersonX is nuts about the Object1.\\
PersonX is smitten with the Object1.\\
PersonX is spellbound by the Object1.\\
PersonX is taken with the Object1.\\
PersonX is wild about the Object1.\\
PersonX loves to drink from a Object1.\\
PersonX would do anything for a Object1.}

\subsubsection{All Paraphrases of positive negative sentences (\texttt{PersonX hates ObjectY})}

\noindent\texttt{PersonX hates Object1.\\ 
PersonX can't stand the Object1.\\ 
PersonX despises the Object1.\\
PersonX detests the Object1.\\
PersonX is annoyed by the Object1.\\
PersonX is bothered by the Object1.\\
PersonX is concerned by the Object1.\\
PersonX is disconcerted by the Object1.\\
PersonX is discouraged by the Object1.\\
PersonX is disgusted by the Object1.\\
PersonX is disheartened by the Object1.\\
PersonX is disquieted by the Object1.\\
PersonX is grieved by the Object1.\\
PersonX is horrified by the Object1.\\
PersonX is irritated by the Object1.\\
PersonX is offended by the Object1.\\
PersonX is pained by the Object1.\\
PersonX is repelled by the Object1.\\
PersonX is revolted by the Object1.\\
PersonX is scandalized by the Object1.\\
PersonX is shocked by the Object1.\\
PersonX is sorrowful by the Object1.\\
PersonX is terrified by the Object1.\\
PersonX is troubled by the Object1.\\
PersonX is vexed by the Object1.\\
PersonX loathes the Object1.\\
The Object1 horrifies PersonX.\\
The Object1 is abhorrent to PersonX.\\
The Object1 nauseates PersonX.\\
The Object1 offends PersonX.\\
The Object1 repulses PersonX.\\
The Object1 revolts PersonX.\\
The Object1 scandalizes PersonX.\\
The Object1 shocks PersonX.\\
The Object1 sickens PersonX.\\
The Object1 terrifies PersonX.\\
The Object1 turns PersonX's stomach.}

\subsection{Structure of Story Structure Robustness Test Sets}
\label{appendix:example-test-sets} 

\subsubsection{Double Room False-Belief Episode}
\noindent\texttt{person1 entered the room1.\\
person2 entered the room1.\\
The object1 is in the container1.\\
The container1 is in the room1.\\
person2 exited the room1.\\
person1 moved the object1 to the container2.\\
The container2 is in the room1.\\
person1 exited the room1.\\
person2 entered the room2.\\
person1 entered the room2.\\
The object2 is in the container3.\\
The container3 is in the room2.\\
person1 exited the room2.\\
person2 moved the object2 to the container4.\\
The container4 is in the room2.\\
person2 exited the room2.
}
\subsubsection{Three Active Characters Story}
\noindent\texttt{person1 entered the room1.\\
person2 entered the room1.\\
person3 entered the room1.\\
The object1 is in the container1.\\
The container1 is in the room1.\\
person2 exited the room1.\\
person1 moved the object1 to the container2.\\
The container2 is in the room1.\\
person1 exited the room1.\\
person3 moved the object1 to the container3.\\
The container3 is in the room1.\\
person3 exited the room1.
}

\subsubsection{True-Belief Interaction, Falsified by Unwitnessed Third-Person Story}
\noindent\texttt{person1 entered the room1.\\
person2 entered the room1.\\
The object1 is in the container1.\\
The container1 is in the room1.\\
person1 moved the object1 to the container2.\\
The container2 is in the room1.\\
person2 exited the room1.\\
person1 exited the room1.\\
person3 entered the room1.\\
person3 moved the object1 to the container1.}

\subsubsection{Four Containers with Multiple Movements}

\noindent\texttt{person1 is in the room1.\\
The object1 is in the container1.\\
The container1 is in the room1.\\
person1 moved the object1 to the container2.\\
The container2 is in the room1.\\
person2 entered the room1.\\
person1 exited the room1.\\
person2 moved the object1 to the container3.\\
The container3 is in the room1.\\
person2 moved the object1 to the container4.\\
The container4 is in the room1.}

\section{Expanded Results}

\paragraph{Experimental Note:} All zero-shot GPT3 (\texttt{text-curie-001} and \texttt{text-davinci-002}) experiments were performed between November 2022 and January 2023. GPT3.5 (\texttt{gpt-3.5-turbo}) and GPT4 (\texttt{gpt-4}) were added in May 2023.

\subsection{Ablating \textsc{filterBasedOnQuestion} from \ourname{}}\label{sec:ablation}

\paragraph{\textsc{filterBasedOnQuestion} definition} This function filters the story $S'$ to obtain an even shorter subset of the original story $S''$ by only keeping edges where at least one of the endpoints represents an entity mentioned in the question.\\

Last step of Algorithm \ref{alg:overview} is applying \textsc{filterBasedOnQuestion}, 
which yields an even shorter story to feed language models. We evaluate the effect this final filter has on the final performances reported by \ourname{}.

\begin{figure}[h]
    \centering
    \includegraphics[width=\linewidth]{images_camera_ready/labels_symbolictom.pdf}
    \includegraphics[width=0.9\linewidth]{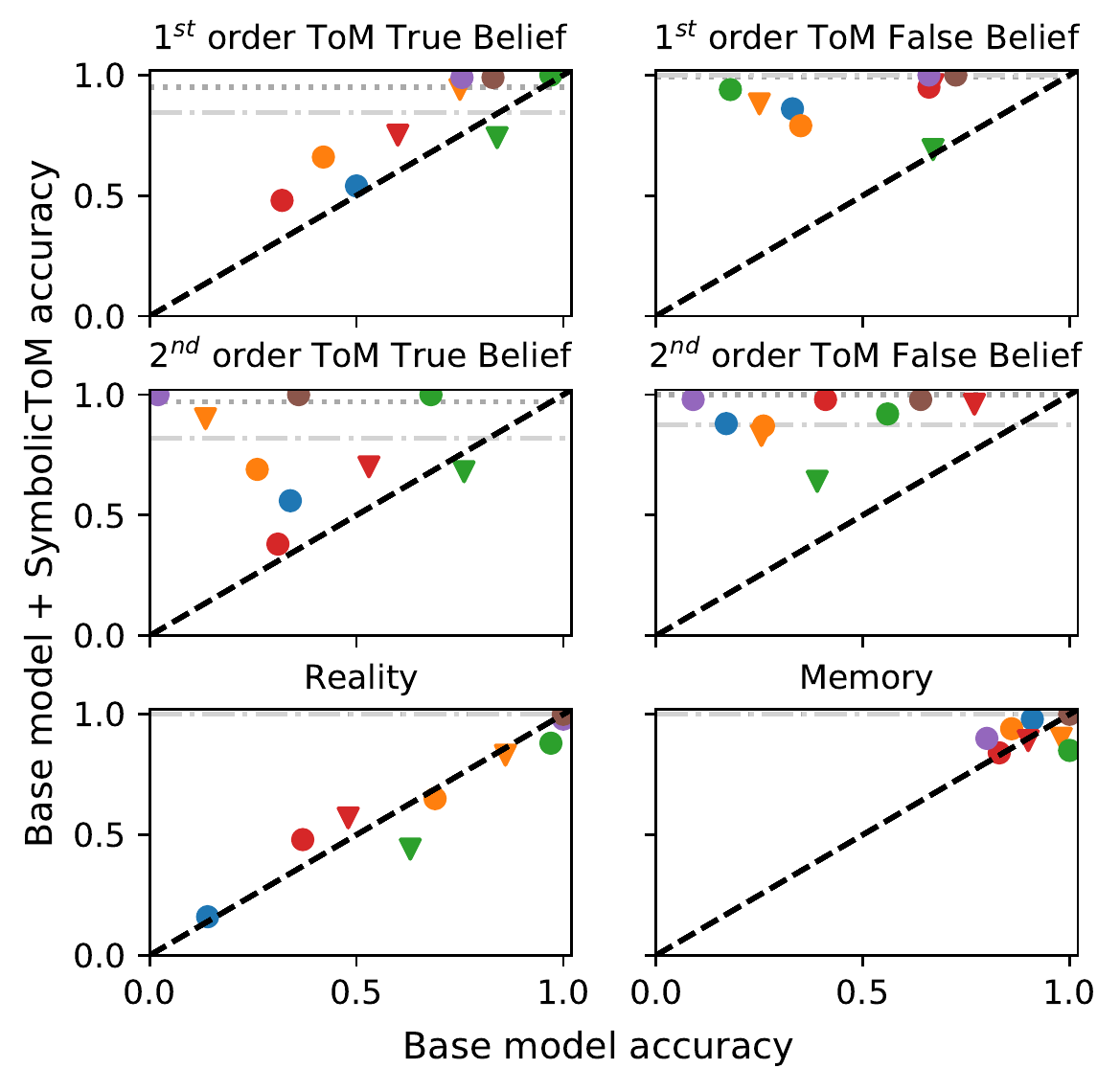}
    \caption{Precision using \ourname{} on ToMi, for several language models without the final filter function. Performance is shown for each question type, dots in upper triangle imply performance improvements. Full results table may be found in Table~\ref{table:perf-regular-test-set-ablation-final-filter}.}
    \label{fig:performance-by-question-ablation-final}
\end{figure}

\textsc{filterBasedOnQuestion} has a positive effect on Macaw-3B, GPT3, Flan-T5-XXL, and LLaMA-7B (+7, +3.5, +12.8, and +15 points in average accuracy gain across all question types), and a mild negative one on Flan-T5-XL, and GPT4 (-5.3, and -4 points of accuracy on average). See Table \ref{table:diff-using-final-filter-or-not} for all differences between executing \ourname{} using this final filtering or not. Figure \ref{fig:performance-by-question-ablation-final} visually represents the accuracy of all models by question type. Regardless of the final filter application, GPT4+\ourname{} significantly outperforms out-of-the-box GPT4 in all four ToM question types and maintains performance on Reality and Memory questions. For Flan-T5-XL, Flan-T5-XL+\ourname{} outperforms Flan-T5-XL significantly in all four ToM question types (e.g. +76 and +36 points in accuracy for first and second-order false belief questions), and shows slight declines for Reality and Memory questions---in line with findings on the full algorithm, but with less stark declines, suggesting that having more entities may help reduce bias towards answering rooms instead of containers. See Table \ref{table:perf-regular-test-set-ablation-final-filter} for the full table of accuracy differences.

Regardless of the final filtering application, \ourname{} shows improvements in theory of mind questions for all models. We only find the filter application to be relevant to beat the base model in theory of mind questions for Flan-T5-XXL.

\subsection{Third-Order Theory of Mind Evaluation}\label{app:third_order_tom}

\begin{table}[h]
\centering
\normalsize
\begin{tabular}{@{}cc@{}}
\toprule
             & \begin{tabular}[c]{@{}c@{}}\textsc{$D_2$'s third-order} \\ \textsc{ToM questions}\end{tabular}             \\ \midrule
\rowcolor{verylightgray} \multicolumn{2}{l}{\textit{Off-the-shelf models}} \\
Macaw-3B     &  13 \\
Flan-T5-XL    &  32 \\
Flan-T5-XXL     &  62 \\
GPT3-Curie   & 28 \\
GPT3-Davinci & 19 \\
GPT3.5 & \phantom{0}8 \\
GPT4 & 26 \\
LLaMA-7B & 22 \\
LLaMA-13B & 39 \\
\midrule
\rowcolor{verylightgray} \multicolumn{2}{l}{\textit{\ourname{} + Off-the-shelf models}} \\
Macaw-3B     & \phantom{0}85 {\small (+72)} \\ 
Flan-T5-XL     & \phantom{0}97 {\small (+65)} \\
Flan-T5-XXL     & \textbf{100} {\small (+38)} \\
GPT3-Curie   & \phantom{0}89 {\small (+61)} \\
GPT3-Davinci & \phantom{0}90 {\small (+71)} \\
GPT3.5 & \textbf{100} \textbf{\small \color{bettergreen} (+91)} \\ 
GPT4 & \textbf{100} {\small (+73)} \\
LLaMA-7B & \phantom{0}90 {\small (+68)} \\
LLaMA-13B &  \phantom{0}95 {\small (+57)} \\
\midrule
\rowcolor{verylightgray} \multicolumn{2}{l}{\textit{Supervised models}} \\
{\small TTT}  & 52 \\
{\small Finetuned GPT3} & 76 \\ \bottomrule
\end{tabular}
\caption{Precision using \ourname{} on all questions from 100 stories for each of the modified test sets $D_i$. Supervised models were trained on ToMi; all others do not require training. Parenthesis reflect differences between using and not using \ourname{}: \textbf{bold} reflects higher overall performance, and {\color{bettergreen} \textbf{green}} reflects the highest net improvements when using \ourname{}. 
}\label{tab:custom-test-sets-third-order}
    \vspace{-0.5em}
\end{table}

We ask all third-order theory of mind questions for each $D_2$ story, such as ``Where does $p_1$ think that $p_2$ thinks that $p_1$ will search for the $o_1$?''. Questions involving $p_2$ will have a final answer $c_1$, since everyone saw $p_2$ leaving. We ask all six possible questions involving $p_2$. We also ask the two third-order theory of mind questions that do not involve $p_2$ nor repeats the same person twice consecutively (``Where does $p_1$ think that $p_3$ thinks that $p_1$ will search for the $o_1$?'' and ``Where does $p_3$ think that $p_1$ thinks that $p_3$ will search for the $o_1$?''), totaling eight questions per $D_2$ story.

Table \ref{tab:custom-test-sets-third-order} shows results for all models using $k=2$ representations (same depth as in the main paper). Using \ourname{} significantly outperforms the supervised baselines and yields dramatic improvements with respect to using the LLMs off-the-shelf. We hypothesize that although the task theoretically requires $k=3$, the second-order theory of mind representation already helps models avoid attending to parts of the story that are inaccessible to relevant characters.

\subsection{Alternative \textsc{resultingState}  Implementations}\label{sec:linguistic-diverisity-using-self-models}
\vspace{-3mm}

\begin{figure}[h]
    \centering

    \includegraphics[width=\linewidth]{images_camera_ready/labels_symbolictom.pdf}
    \includegraphics[width=\linewidth]{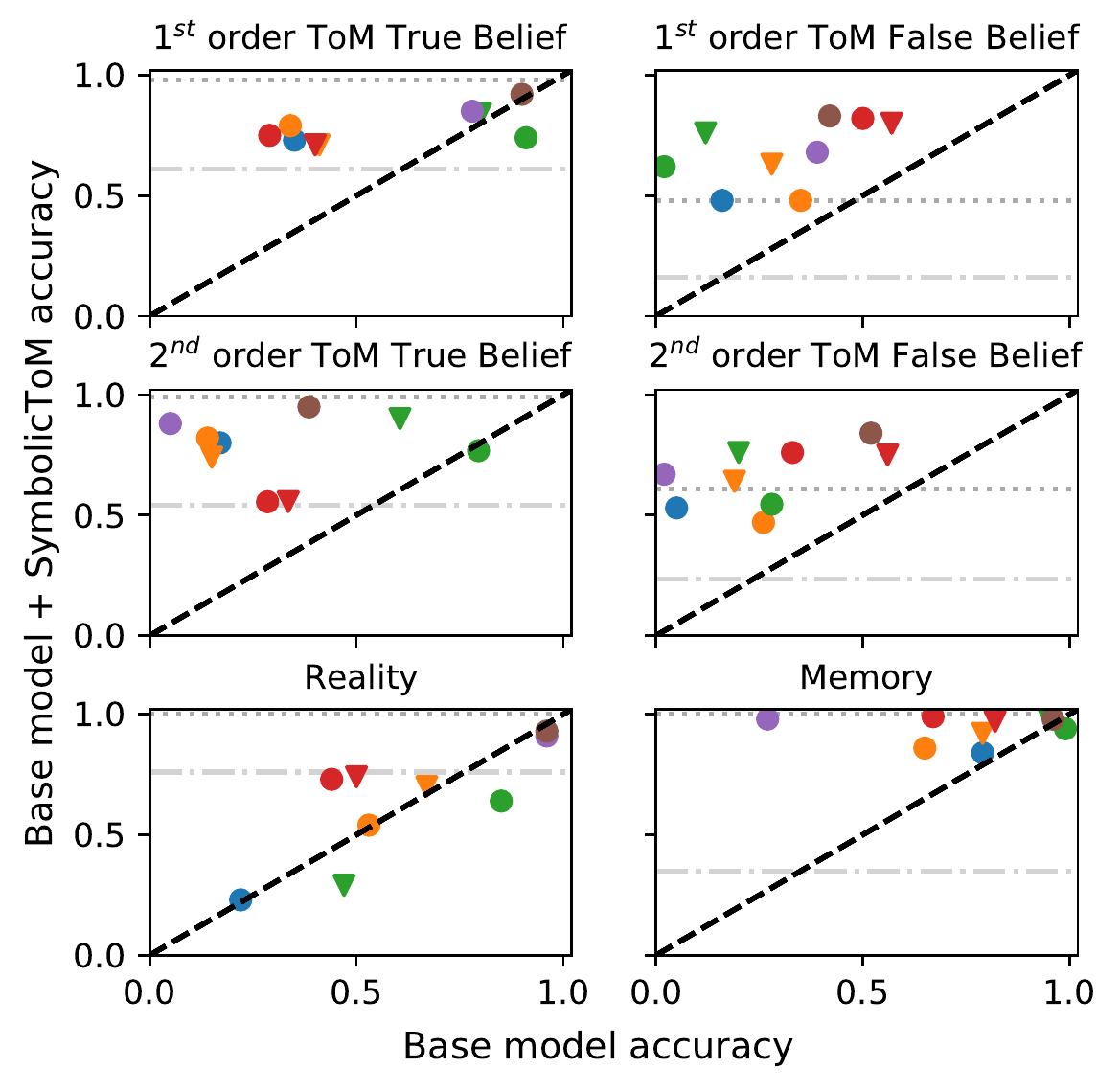}

    \caption{Results for ParaphrasedToMi when using the same model for implementing the \textsc{resultingState} function as in the final question-answering task (except using Davinci for Macaw, who did not show reliable enough few shot-prompting). Dots in upper triangle imply performance with \ourname{} is higher than using the base model out-of-the-box. Horizontal lines reflect supervised models' performance (higher is\,better).}

    \label{fig:use-self-model-for-resulting-state}
\end{figure}

\textsc{resultingState($s$)} refers to the state of the world after $s$ has been performed. For example, if ``Oliver moved the apple to the box'', then the resulting state is that ``The apple is in the box''. If ``Oliver exited the bedroom'', the resulting state would be that ``Oliver is no longer in the bedroom''. These are the relationships that we may insert in a context graph---actions are instantaneous and do not reflect an observable state.\\ 

In this section, we explore using the same LLM for implementing \textsc{resultingState} as well as the final inference. In the main text, we use Davinci for all non-GPT3-based models.\\

\vspace{-1mm}
We find GPT3 to be among the most reliable to answer the resulting state of a given action in a zero-shot (Davinci) or two-shot (Curie) manner. Similarly, GPT3.5 and GPT4 perform well zero-shot: for experiments, we use GPT3.5 zero-shot and GPT4 two-shot to improve the resulting phrasing stability.

Additional exploration shows that although Flan-T5 models perform worse zero-shot than GPT models, they are capable of performing this task with more careful prompting. Figure \ref{fig:use-self-model-for-resulting-state} shows the results after nine-shot prompting Flan-T5-XL and eleven-shot prompting Flan-T5-XXL. Our explorations show that LLaMA models require fewer demonstrations than the Flan-T5 models to compute the resulting state: we observe highly reliable results when using six-shot prompting for LLaMA-7B, and seven-shot prompting for LLaMA-13B. Accuracy using LLaMA was even higher than when using GPT3.

\subsection{Detailed Result Tables}

All results in the appendix show accuracy as a ratio (between 0 and 1). For simplicity of reading, in the main text, they are referred to in percentages (values 0 to 100, higher is better). Figures \ref{table:perf-regular-test-set}, \ref{table:perf-regular-test-set-ablation-final-filter}, and \ref{table:diff-using-final-filter-or-not} show performances when applying the final filtering function, when not applying it, and the difference in performance between the two,\,respectively.

\begin{table*}[t]
\centering
\begin{tabular}{@{}ccccccc@{}}
\toprule
         & 1st TB          & 1st FB          & 2nd TB          & 2nd FB          & Reality         & Memory          \\ \hline \midrule
Macaw-3B       & 0.86 {[}0.50{]} & 0.79 {[}0.33{]} & 0.86 {[}0.34{]} & 0.84 {[}0.17{]} & 0.10 {[}0.14{]} & 0.95 {[}0.91{]} \\
GPT3-Curie     & 0.77 {[}0.42{]} & 0.82 {[}0.35{]} & 0.73 {[}0.26{]} & 0.89 {[}0.26{]} & 0.61 {[}0.69{]} & 0.99 {[}0.86{]} \\
GPT3-Davinci         &                               0.96 {[}0.75{]} & 0.96 {[}0.25{]} &  0.93 {[}0.14{]}  & 0.90 {[}0.26{]} & 0.77 {[}0.86{]} & 0.98 {[}0.98{]} \\
Flan-T5-XL      & 0.98 {[}0.97{]} & 0.80 {[}0.18{]} & 0.98 {[}0.68{]} & 0.78 {[}0.56{]} & 0.73 {[}0.97{]} & 1.00 {[}1.00{]} \\
Flan-T5-XXL     & 0.98 {[}0.84{]} & 0.95 {[}0.67{]} & 1.00 {[}0.76{]} & 0.90 {[}0.39{]} & 0.13 {[}0.63{]} & 1.00 {[}1.00{]} \\ 
LLaMA-7B      &                                      0.82 {[}0.32{]} &  0.95 {[}0.66{]} &  0.66 {[}0.31{]} &  0.72 {[}0.41{]} &  0.87 {[}0.37{]} &  1.00 {[}0.83{]} \\
LLaMA-13B      &                                     0.82 {[}0.60{]} &  0.86 {[}0.67{]} &  0.70 {[}0.53{]} &  0.62 {[}0.77{]} &  0.87 {[}0.48{]} &  1.00 {[}0.90{]} \\
GPT3.5         &                                     0.97 {[}0.76{]} &  0.95 {[}0.66{]} &  0.99 {[}0.02{]} &  0.87 {[}0.09{]} &  0.98 {[}1.00{]} &  0.99 {[}0.80{]} \\
GPT4            &                                    0.98 {[}0.83{]} &  0.94 {[}0.73{]} &  0.98 {[}0.36{]} &  0.89 {[}0.64{]} &  0.94 {[}1.00{]} &  1.00 {[}1.00{]} \\ \midrule
Finetuned GPT3 & 0.95            & 0.99            & 0.97            & 1.00            & 1.00            & 1.00            \\
TTT-learned                      & 0.84            & 1.00            & 0.82            & 0.88            & 1.00            & 1.00            \\ \bottomrule
\end{tabular}
\caption{Performance per model and question using \ourname{}, with out-of-the-box performance shown in brackets (100 samples per question type). Bottom rows represent supervised baselines.}\label{table:perf-regular-test-set}
\end{table*}


\begin{table*}[t]
\centering
\begin{tabular}{@{}ccccccc@{}}
\toprule
                                                   & 1st TB          & 1st FB          & 2nd TB          & 2nd FB          & Reality         & Memory          \\ \midrule
Macaw-3B                                           & 0.54 {[}0.50{]} & 0.86 {[}0.33{]} & 0.56 {[}0.34{]} & 0.88 {[}0.17{]} & 0.16 {[}0.14{]} & 0.98 {[}0.91{]} \\
GPT3-Curie                                         & 0.66 {[}0.42{]} & 0.79 {[}0.35{]} & 0.69 {[}0.26{]} & 0.87 {[}0.26{]} & 0.65 {[}0.69{]} & 0.94 {[}0.86{]} \\

GPT3-Davinci                  &                      0.94 {[}0.75{]} &  0.88 {[}0.25{]} &  0.90 {[}0.14{]} &  0.83 {[}0.26{]} &  0.83 {[}0.86{]} &  0.90 {[}0.98{]} \\

Flan-T5-XL                                         & 1.00 {[}0.97{]} & 0.94 {[}0.18{]} & 1.00 {[}0.68{]} & 0.92 {[}0.56{]} & 0.88 {[}0.97{]} & 0.85 {[}1.00{]} \\
Flan-T5-XXL                                        & 0.74 {[}0.84{]} & 0.69 {[}0.67{]} & 0.68 {[}0.76{]} & 0.64 {[}0.39{]} & 0.44 {[}0.63{]} & 1.00 {[}1.00{]}   \\

LLaMA-7B  &                                          0.48 {[}0.32{]} & 0.95 {[}0.66{]} & 0.38 {[}0.31{]} & 0.98 {[}0.41{]} & 0.48 {[}0.37{]} & 0.84 {[}0.83{]} \\
LLaMA-13B    &                                         0.75 {[}0.60{]} & 0.96 {[}0.67{]} & 0.70 {[}0.53{]} & 0.96 {[}0.77{]} & 0.57 {[}0.48{]} & 0.89 {[}0.90{]}   \\
GPT3.5    &                                           0.99 {[}0.76{]} & 1.00 {[}0.66{]} & 1.00 {[}0.02{]} & 0.98 {[}0.09{]} & 0.98 {[}1.00{]} & 0.90 {[}0.80{]}   \\
GPT4    &                                            0.99 {[}0.83{]} & 1.00 {[}0.73{]} & 1.00 {[}0.36{]} & 0.98 {[}0.64{]} & 1.00 {[}1.00{]} & 1.00 {[}1.00{]} \\ \midrule
Finetuned GPT3 & 0.95            & 0.99            & 0.97            & 1.00            & 1.00            & 1.00            \\
TTT-learned                     & 0.84            & 1.00            & 0.82            & 0.88            & 1.00            & 1.00            \\ \bottomrule
\end{tabular}
\caption{Performance per model and question using \ourname{} without \textsc{filterBasedOnQuestion}, with out-of-the-box performance shown in brackets (100 samples per question type). Bottom rows represent supervised baselines.}\label{table:perf-regular-test-set-ablation-final-filter}
\end{table*}

\begin{table*}[]
\centering
\begin{tabular}{@{}ccccccc@{}}
\toprule
\multicolumn{1}{l}{} & 1st TB & 1st FB & 2nd TB & 2nd FB & Reality & Memory \\ \midrule
Macaw-3B             & 0.32   & -0.07  & 0.30   & -0.04  & -0.06   & -0.03  \\
GPT3-Curie           & 0.11   & 0.03   & 0.04   & 0.02   & -0.04   & 0.05   \\
GPT3-Davinci         & 0.02   & 0.08   & 0.03   & 0.07   & -0.06   & 0.08   \\
Flan-T5-XL           & -0.02  & -0.14  & -0.02  & -0.14  & -0.15   & 0.15   \\
Flan-T5-XXL          & 0.24   & 0.26   & 0.32   & 0.26   & -0.31   & 0.00   \\
LLaMA-7B             & 0.34   & 0.00   & 0.28   & -0.26  & 0.39    & 0.16   \\
LLaMA-13B            & 0.07   & -0.10  & 0.00   & -0.34  & 0.30    & 0.11   \\
GPT3.5               & -0.02  & -0.05  & -0.01  & -0.11  & 0.00    & 0.09   \\
GPT4                 & -0.01  & -0.06  & -0.02  & -0.09  & -0.06   & 0.00   \\ \bottomrule
\end{tabular}
\caption{Differences between accuracy of base models using \ourname{} with the final \textsc{filterBasedOnQuestion} filter, and without using the final filter. As shown in Table \ref{table:perf-regular-test-set} and \ref{table:perf-regular-test-set-ablation-final-filter}, both versions are still far superior to not using \ourname{}.}\label{table:diff-using-final-filter-or-not}
\end{table*}

\end{document}